\documentclass[german,english]{bvm2015}

\title{The Random Forest Classifier in \mbox{WEKA}: Discussion and New Developments for Imbalanced Data}
\titlerunning{Balanced Random Forest Classifier in \mbox{WEKA}}

\subtitle{}

\author{Mario Amrehn$^1$, Firas Mualla$^1$, Elli Angelopoulou$^1$, Stefan Steidl$^1$ and \newline Andreas Maier$^{1,2}$}

\authorrunning{Mario Amrehn et al.}

\institute{%
$^1$Pattern Recognition Lab, Friedrich-Alexander University\\ Erlangen-N{\"u}rnberg \mbox{(FAU)}, Germany\\
$^2$Erlangen Graduate School in Advanced Optical Technologies \mbox{(SAOT)}, Germany%
}

\email{mario.amrehn@fau.de}

\usepackage{listings}

\begin{document}

%
\selectlanguage{english}

\maketitle
\begin{abstract}
Data analysis and machine learning have become an integrative part of the modern scientific methodology, 
providing automated techniques to predict further information based on observations. 
One of these classification and regression techniques is the random forest approach.
Those decision tree based predictors are best known for their good computational performance and scalability.
However, in case of severely imbalanced training data, as often seen in medical studies' data with large control groups, 
the training algorithm or the sampling process has to be altered in order to improve the prediction quality for minority classes.
In this work, a balanced random forest approach for \mbox{WEKA} is proposed.
Furthermore, the prediction quality of the unmodified random forest implementation and the new balanced random forest version for \mbox{WEKA} are evaluated against reference implementations in R.
Two-class problems on balanced data sets and imbalanced medical studies' data are investigated.
A superior prediction quality using the proposed method for imbalanced data is shown compared to the other three techniques.
\end{abstract}%

\section{Introduction}%

In computer aided decision systems, machine learning is used to help a physician in diagnosing a patient.
One of the most common functions performed by such systems is the classification task where a label is assigned to each patient (query case) based on a defined number of clinical findings.
The label represents the patients' membership in one of predefined classes representing possible diagnoses.
Those systems have been applied for the diagnosis of various diseases, such as hypothyroidism and cancer. %
Commonly produced data by such investigations are predominately composed of negative or healthy samples with only a small percentage of positive or diseased ones,
leading to class imbalance problems with suboptimal classification performance.
In class imbalance problems, inputting all the data into the classifier to build up the learning model will usually lead to a learning biased towards the majority class, 
optimizing the overall accuracy without considering the relative distribution of each class. %
The class imbalance problem is almost ubiquitous in real world data, causing trouble to a large segment of the machine learning community.

\noindent One of the most popular frameworks used for classification in machine learning is the Waikato Environment for Knowledge Analysis project (\mbox{WEKA})\,\cite{hall2009weka}.
Initially funded by the New Zealand government, \mbox{WEKA} has a widespread acceptance in both academia and business with an active community and more than 6.5 million downloaded copies since the year 2000.
It provides a unified workbench for an easy access to state-of-the-art techniques in machine learning.
However, the built-in random forest (RF) classification approach does not handle imbalanced data well.
The contribution of this work is extending the \mbox{WEKA} data mining framework with a RF classifier which is able to deal with imbalanced data sets.

\section{Materials and Methods}%

The random forest classification and regression method induces each constituent decision tree from a bootstrap sample of the training data \cite{breiman2001random}.
Dealing with severely imbalanced data, there is a significant probability that a sample contains few or none of the minority class. This input results in a tree with poor performance for predicting this class \cite{chen2004using}. %
Simple up-sampling would render cross-validation and out-of-bag error measures meaningless due to duplicates from the training set in the holdout set.
Down-sampling severely imbalanced data before classification however, would lead to most instances of the majority class not used during training. %
Further research on extremely imbalanced data sets by Ling and Li\,\cite{ling1998data} and Drummond et al.\,\cite{drummond2003c4} 
shows that the classification quality with respect to a performance measure of the minority class 
can be improved either by down-sampling the majority class or over-sampling the minority class using a tree based approach. %
According to Kubat and Matwin\,\cite{kubat1997addressing}, down-sampling has the edge over over-sampling. %
Chen et al.\,\cite{chen2004using} conclude, that both balanced random forest (\mbox{BRF}) and weighted random forest have superior performance compared to a non-balancing approach. %
The learning process of the original \mbox{WEKA} \mbox{RF} classifier operates in four steps: 
\begin{enumerate}%
 \item Bootstrap samples $\mathbf{B}_i$ for every tree $t_i$ are drawn by randomly selecting instances with replacement from $\mathbf{X}$ until the sizes of $\mathbf{B}_i$ and $\mathbf{X}$ are equal. 
 \item A random subset of features are selected for each $\mathbf{B}_i$ and used for the training of tree $t_i$ in the forest.
 \item An information gain metric is used to grow unpruned decision trees.
 \item The final classification result is the most common of the individual tree predictions.
\end{enumerate}%

Due to the bootstrap sampling with replacement (which is important to avoid over-fitting), every $\mathbf{B}_i$ is likely to include duplicates.
When drawing with replacement $n$ values out of a large set of $n$ unique and equally likely elements, 
the expected number of unique draws is $n\cdot(1 - 1 / e)$ \cite{aslam2007estimating}, which is around $63.21\%$.
In severely imbalanced data, the holdout of more than a third of the original training data is likely to lead to some $t_i$ trained on only a few or not even a single instance of the minority class, depending on $\mathbf{s}$ and the size of $\mathbf{X}$. 
The proposed balanced random forest approach for \mbox{WEKA} alters the step of bootstrap sample generation for each tree as described in \cite{chen2004using}.

\subsection*{Implementation}%

With \mbox{BRF}, the distribution of classes in the training bootstrap sample $\mathbf{B}_i \subset \mathbf{X}$ for each random decision tree $t_i$ is altered, 
where $\mathbf{X}$ is the training data. The sample size is denoted as $\mathbf{s}$ consisting of $\mathbf{s}_c$ for classes $c$ in $\mathbf{X}$.
In \mbox{RF} $\mathbf{s}$ is set to the distribution of classes in $\mathbf{X}$.
In this analysis, each element of $\mathbf{s}$ is set to the number of training data for the minority class.
The balancing process and training of the \mbox{BRF} classifier using under-sampling in R is denoted as follows: %
{\small\begin{lstlisting}
sampsize <- rep(min(as.integer(summary(train\$myClass))), 2)
brf <- randomForest(myClass ~ .,
   data=train, ntree=ntree, importance=TRUE,
   sampsize=sampsize
)
\end{lstlisting}}

With the current \mbox{WEKA} data mining toolkit in developer version 3.7.12, there is no option to edit the sample size as shown in R.
Instead, the elements of $\mathbf{s}$ are always equal to the distribution of classes in the entire training set $\mathbf{X}$. 
The proposed extension of the random forest classification method provides an addition to the Java interface in \mbox{WEKA} to set the sample size during bagging:
{\small\begin{lstlisting}
BalancedRandomForest cl = new BalancedRandomForest();
cl.setNumTrees(ntree);
cl.setSampleSize(sampsize); // new in BRF
cl.buildClassifier(train);
\end{lstlisting}}

We added two classes to the \mbox{WEKA} framework implementation in order to include prediction quality balancing for the minority class. %
\texttt{BalancedBagging} 
extends \texttt{Bagging} and includes the \texttt{sampleSize} member variable.
If set, \\\mbox{\texttt{get\-TrainingSet(int)}} picks samples at random with replacement from each class' data according to its sample size value.
\texttt{BalancedRandomForest} 
extends base class \texttt{RandomForest} with a method \texttt{setSampleSize(int[])} and overrides the \texttt{buildClassifier(Instances)} method in order to use the modified sampling when generating the training set $\mathbf{X}^\mathbf{s}_i$ for a tree $t_i$. %
\mbox{WEKA} in its default configuration uses means/modes for the imputation of missing information in a data set \cite{schafer1998multiple}.
Since \mbox{WEKA} version 3.7.0 expectation-maximization imputation is also possible. The \texttt{randomForest} package for R provides two ways 
to impute missing values in a data set in accordance with the original descriptions of RF by Breiman and Cutler\,\cite{breiman2001random}.
The first method \texttt{na.roughfix} fills in the column's median value for continuous variables and the most frequent non-missing 
discrete value for classes, breaking ties at random.
The second method fills in missing values using the first method's imputed values, then trains a random forest with the completed data.
The proximity matrix from the random forest is used to update the imputed values. Finally, the model is trained using the RF-imputed data set.
For increased comparability of the classification results, here the simple imputation methods of both frameworks are used.

\subsection*{Data Sets}%

For the evaluation, three data sets with severely imbalanced class distributions are used as well as two balanced data sets.
Only large data sets are chosen to reduce the influence of outliers in the evaluation and therefore ranking of the methods concerned. \textbf{Tab.}\,\ref{0000-summaryDatasets} contains a summary of the data provided.
All sets have exactly two distinct class labels. %
The medical conditions concerned by the data sets are difficult to diagnose in an early state without an optimized computer aided decision system.
A proper classification result is therefore crucial for the patient.
The features of the \texttt{HY1} and \texttt{HY2} data sets represent biometric properties from more than $3100$ individuals each, 
classified into healthy and hypothyroid.
Hypothyroidism is a disorder of the endocrine system in which the thyroid gland does not produce enough thyroid hormones.
In children, this disease may lead to cretinism, a severe delay in growth and intellectual development. %
The \texttt{ESS} data set holds information about $3163$ subjects' diagnoses regarding the euthyroid sick syndrome, a state of adaptation or dysregulation of thyrotropic feedback control.
This condition is often seen in starvation, critical illness or patients in intensive care units.
The studies used as balanced data sets for the evaluation are the popular database for the classification of spam and non-spam in text messages by Mark Hopkins et al.\ and a repository of the chess end-game \mbox{King+Rook} versus \mbox{King+Pawn} on \mbox{a7} by Alen Shapiro\,\cite{zdrahal1981recognition}, 
used for a prediction whether white can win after moving as described in the attributes.
Both data sets have a rather balanced ratio of class frequencies.
\begin{table}
\begin{tabular*}{\textwidth}{@{\extracolsep{\fill}} lcccc}%
Dataset & Imbalanced & Minority class & Num.\ Features & Num.\ Instances\\\hline%
HY1 & yes & $4.77\%$ & $25$ & $3,163$ \\
HY2 & yes & $6.12\%$ & $29$ & $3,772$ \\ 
ESS & yes & $9.26\%$ & $25$ & $3,163$ \\
SPAM & no & $39.40\%$ & $57$ & $4,601$ \\
EG & no & $47.78\%$ & $36$ & $3,196$ \\
\end{tabular*} %
\caption{Imbalanced and balanced data sets with two classes used for the evaluation of the new balanced random forest in \mbox{WEKA}.
Sources: (HY1|ESS) http://hakank.org/weka/(HY|SE).arff,
(HY2|SPAM|EG) http://repository.seasr.org/Datasets/UCI/arff/(sick|spambase|kr-vs-kp).arff.}%
\label{0000-summaryDatasets}%
\end{table}%

\section{Results}%

\begin{figure}
	\centering%
	\begin{tabular*}{\textwidth}{@{\extracolsep{\fill}} cc}%
		Imbalanced data (HY1, HY2, ESS) & Balanced data (SPAM, EG) \\
		\includegraphics[trim={8 11 7.5 8},clip,width=0.49\textwidth]{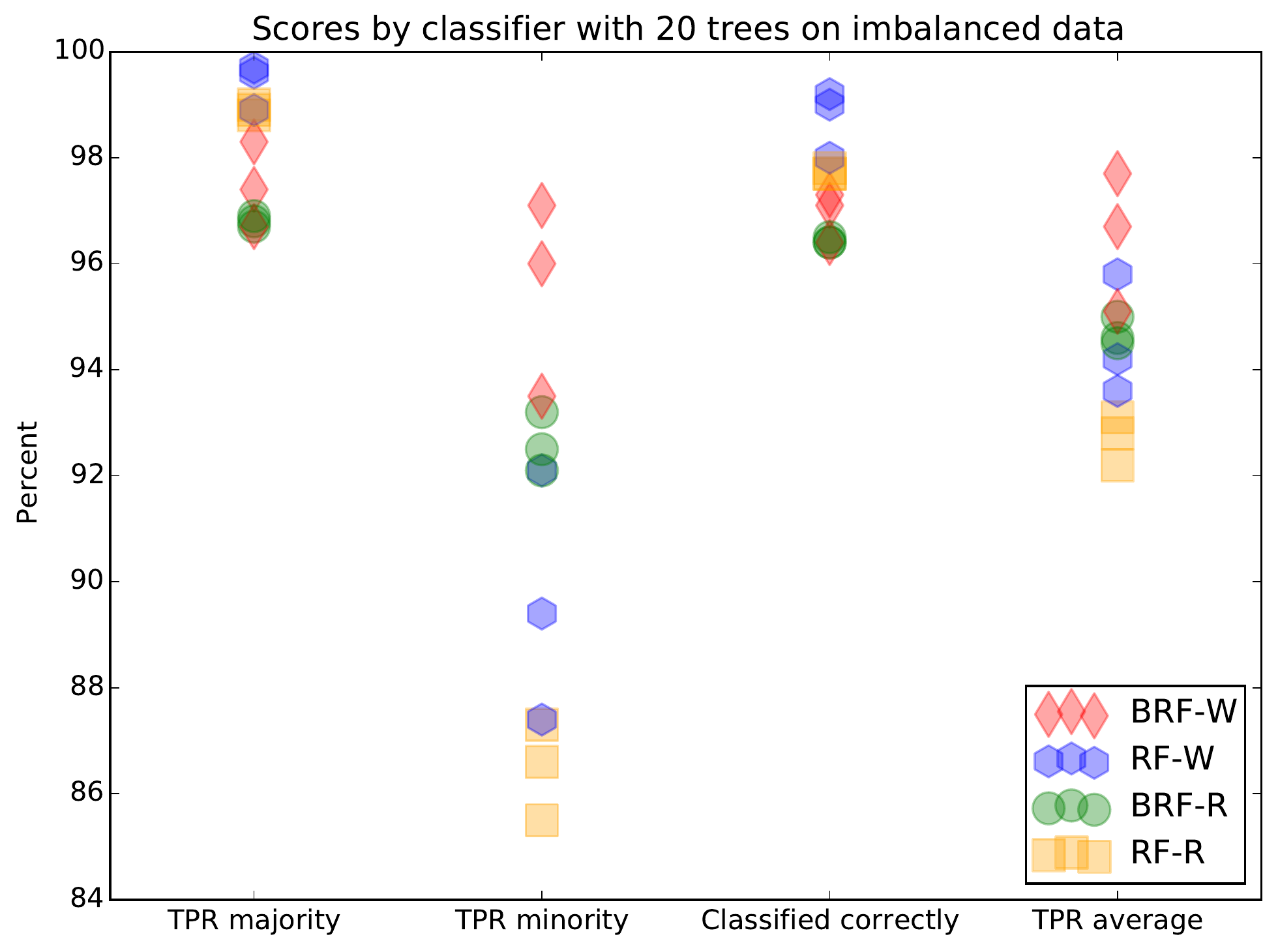} & %
		\includegraphics[trim={8 11 7.5 8},clip,width=0.49\textwidth]{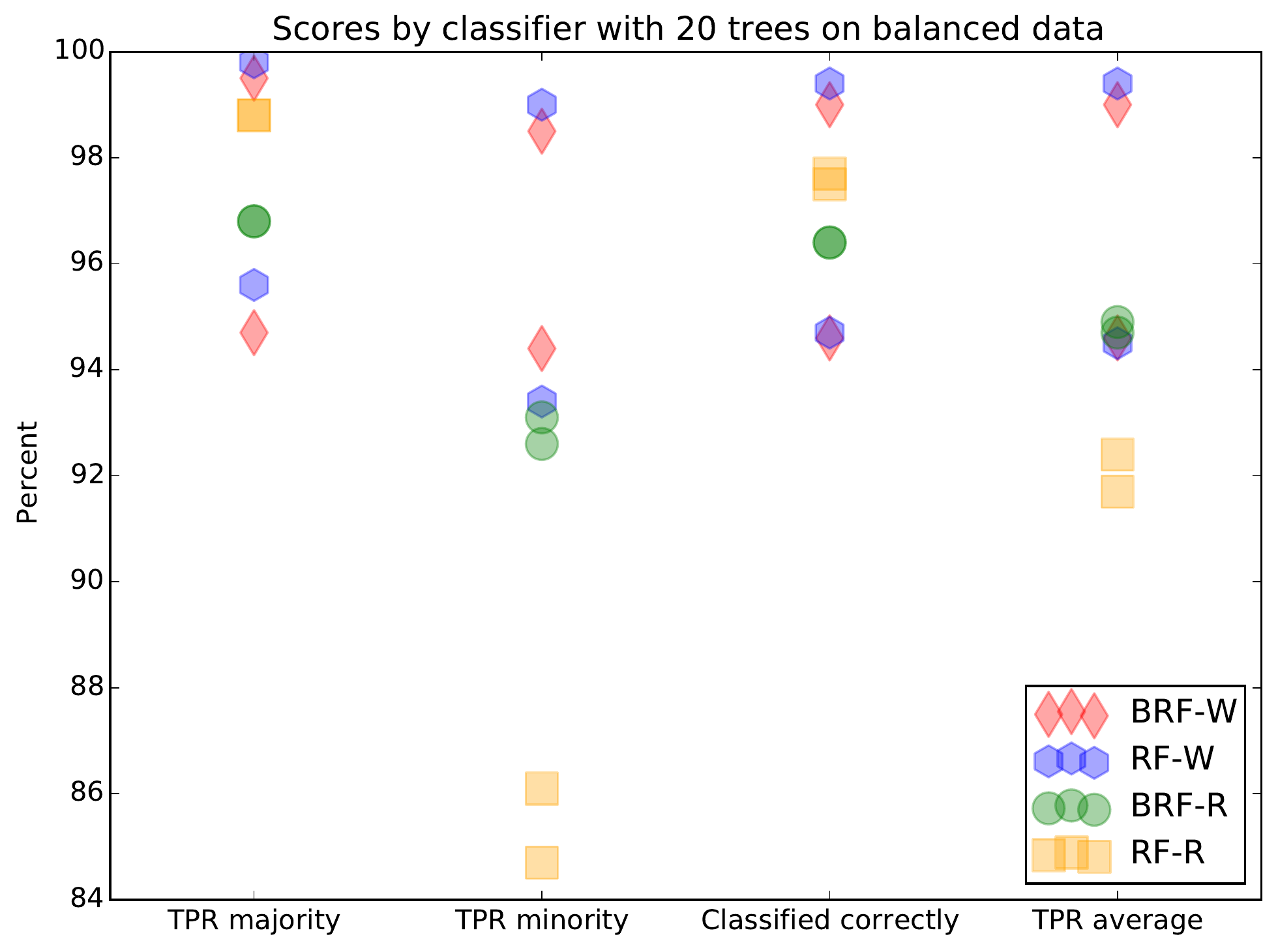} \\ %
		\includegraphics[trim={8 11 7.5 8},clip,width=0.49\textwidth]{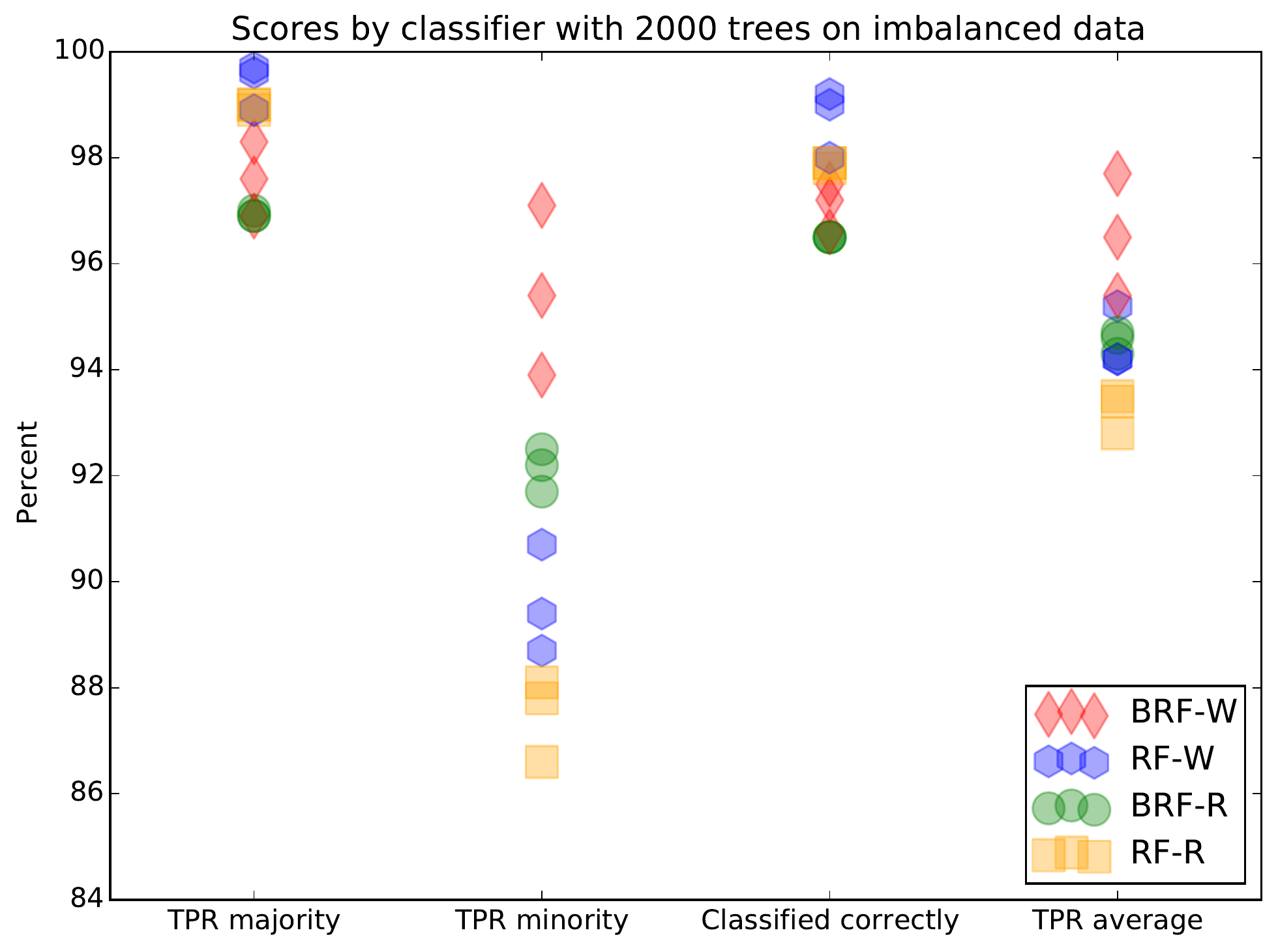} & %
		\includegraphics[trim={8 11 7.5 8},clip,width=0.49\textwidth]{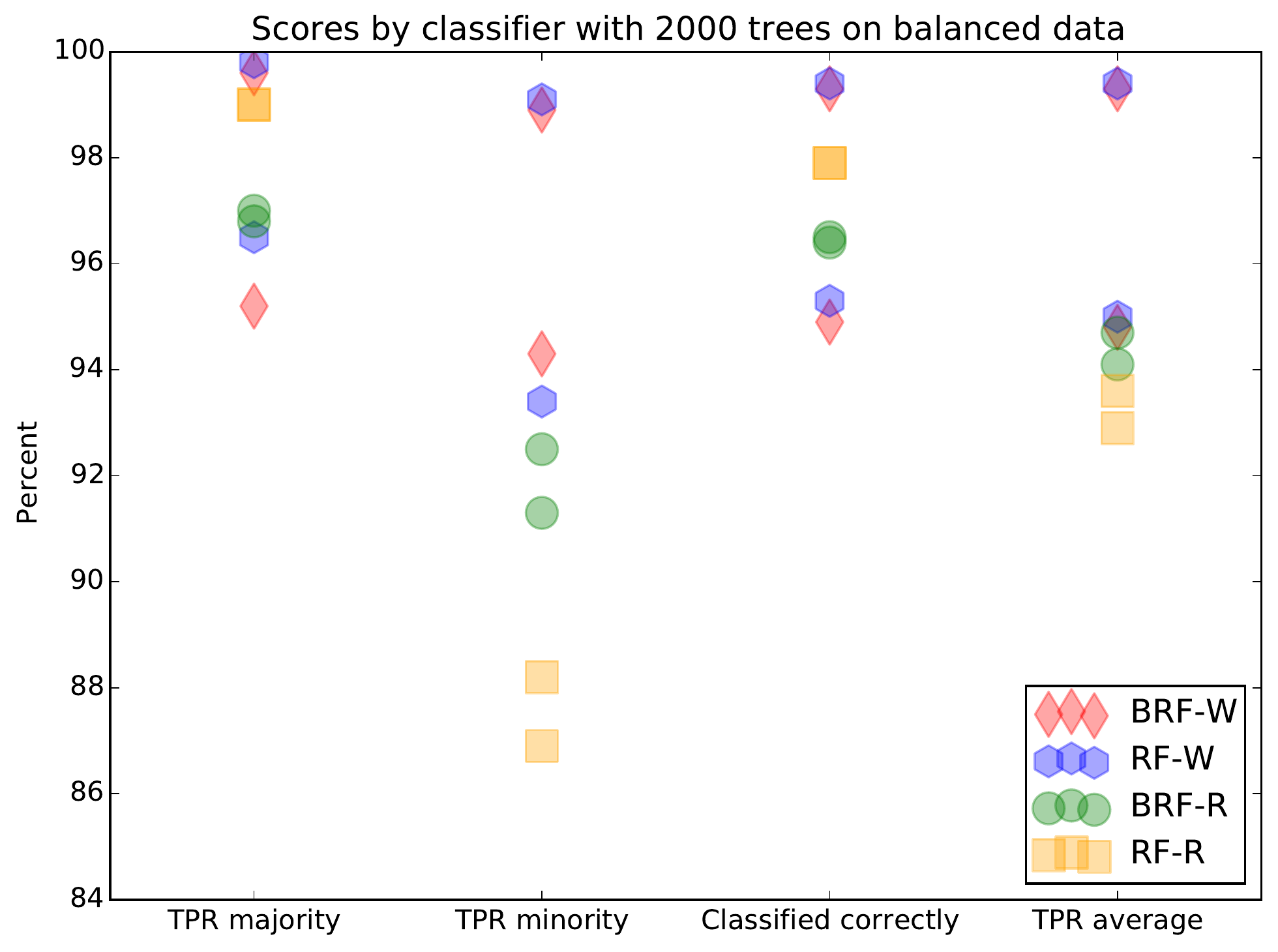} \\ %
	\end{tabular*}%
	\caption{Results are obtained by ten-fold cross-validation.
		\mbox{RFs} in the upper row are generated from $20$ trees, the lower row utilizes \mbox{RFs} with $2,000$ trees.%
	}%
	\label{0000-evaluationResultsFig}
\end{figure}

For the evaluation, the data presented in \textbf{Tab.}\,\ref{0000-summaryDatasets} are used to train \mbox{WEKA's} default random forest classifier (RF-W), 
the proposed balanced \mbox{RF} classifier (\mbox{BRF-W}) based on \mbox{RF-W}, as well as models from \mbox{R's} \textit{randomForest} package with (mox{BRF-R}) and without (\mbox{RF-R}) setting its \texttt{sampsize} parameter.
A ten-fold cross-validation is performed for a classification quality comparison of the methods introduced.
Since the evaluation is dependent on the number of trees as a classifier parameter, each evaluation is repeated with $20$, $100$ and $2,000$ trees in \textbf{Fig.}\,\ref{0000-evaluationResultsFig} and \textbf{Tab.}\,\ref{0000-summaryResults} respectively.
$\text{TPR}_\text{avg}$ measures classification quality of imbalanced data.

\begin{table}
\centering%
\begin{tabular*}{\textwidth}{@{\extracolsep{\fill}} llllllllllll}%
Trees& & & $100$ & & & & & $2,000$ \\
Data & System 	& & $\text{TPR}_\text{ma}$ 	& $\text{TPR}_\text{mi}$ & CCR	& $\text{TPR}_\text{avg}$ &  & $\text{TPR}_\text{ma}$ 	& $\text{TPR}_\text{mi}$ 		& CCR	& $\text{TPR}_\text{avg}$ \\\hline
HY1 & BRF-W     &	&			97.5 & 95.4 & 97.4 & 96.4        &  &           97.6 & 95.4 & 97.5 & 96.5 \\
   & RF-W      	&	&			99.6 & 91.4 & 99.2 & 95.5                 &  &           99.6 & 90.7 & 99.2 & 95.2 \\
   & BRF-R      & & 		96.8 & 92.6 & 96.4 & 94.7                 &  &           96.9 & 91.7 & 96.5 & 94.3 \\
   & RF-R       & & 		98.9 & 88.1 & 97.8 & 93.5                 &  &           98.9 & 86.6 & 97.8 & 92.8 \\\hline
HY2 & BRF-W     &	&			98.3 & 97.0 & 97.0 & 97.6        &  &           98.3 & 97.1 & 97.2 & 97.7 \\
     & RF-W     &	&			99.7 & 88.3 & 99.0 & 94.0                 &  &           99.7 & 88.7 & 99.0 & 94.2 \\
     & BRF-R    &	&			97.0 & 92.9 & 96.6 & 94.9                 &  &           97.0 & 92.2 & 96.5 & 94.6 \\
     & RF-R     &	&			98.8 & 86.9 & 97.7 & 92.9                 &  &           99.0 & 88.1 & 97.9 & 93.5 \\\hline
ESS & BRF-W     &	&			96.9 & 93.5 & 96.6 & 95.2        &  &           96.9 & 93.9 & 96.6 & 95.4 \\
   & RF-W      	&	&			98.9 & 90.4 & 98.1 & 94.7                 &  &           98.9 & 89.4 & 98.0 & 94.2 \\
   & BRF-R      &	&			96.8 & 92.9 & 96.5 & 94.8                 &  &           96.9 & 92.5 & 96.5 & 94.7 \\
   & RF-R       &	&			98.9 & 86.8 & 97.8 & 92.8                 &  &           99.0 & 87.8 & 97.9 & 93.4 \\\hline
SPAM & BRF-W 		&	&			95.2 & 94.3 & 94.8 & 94.7                 &  &           95.2 & 94.3 & 94.9 & 94.8 \\
     & RF-W			&	&			96.4 & 93.5 & 95.3 & 95.0        &  &           96.5 & 93.4 & 95.3 & 95.0 \\
     & BRF-R		&	&			96.9 & 91.6 & 96.4 & 94.2                 &  &           96.8 & 92.5 & 96.4 & 94.7 \\
     & RF-R			&	&			99.0 & 87.0 & 97.9 & 93.0                 &  &           99.0 & 86.9 & 97.9 & 92.9 \\\hline
EG & BRF-W 			&	&			99.6 & 98.8 & 99.2 & 99.2                 &  &           99.6 & 98.9 & 99.3 & 99.3 \\
	 & RF-W				&	&			99.8 & 99.1 & 99.5 & 99.5        &  &           99.8 & 99.1 & 99.4 & 99.4 \\
   & BRF-R			&	&			97.0 & 92.7 & 96.6 & 94.8                 &  &           97.0 & 91.3 & 96.5 & 94.1 \\
   & RF-R				&	&			99.0 & 87.1 & 97.9 & 93.0                 &  &           99.0 & 88.2 & 97.9 & 93.6 \\
\end{tabular*}%
\caption{Results of the ten-fold cross-validation with $100$ and $2,000$ trees.
	The true positive rates for the majority $\text{TPR}_\text{ma}$ and minority class $\text{TPR}_\text{mi}$, percentage of cases correctly classified (CCR) and mean of the two true positive rates $\text{TPR}_\text{avg}$ are presented in percent values.%
}%
\label{0000-summaryResults}%
\end{table}%

\section{Discussion}%

The performed cross-validation indicates a significant improvement in prediction quality for imbalanced data sets in regard to the true positive rate of the minority class as well as the class averaged true positive rate. The $\text{TP}_\text{ma}$ decreases slightly, 
due to the decision rules' redistribution of attribute importance during the split criterion determination in each tree.
A comparative evaluation between the balanced random forest implementation in R and the new \mbox{BRF} implemented in \mbox{WEKA} was presented.
Balancing the data during bootstrap sample generation yields an improvement in the prediction error of the minority class in severely imbalanced data sets.
In this regard, the proposed balanced random forest classifier in \mbox{WEKA} is superior to its original unbalanced version.

%

\bibliographystyle{bvm2015}


\begin{thebibliography}{1}
	
	\bibitem{hall2009weka}
	Hall M, Frank E, Holmes G, Pfahringer B, Reutemann P, Witten IH.
	\newblock The WEKA data mining software: an update.
	\newblock SIGKDD explorations newsletter. 2009;11(1):10--18.
	
	\bibitem{breiman2001random}
	Breiman L.
	\newblock Random forests.
	\newblock Machine learning. 2001;45(1):5--32.
	
	\bibitem{chen2004using}
	Chen C, Liaw A, Breiman L.
	\newblock Using random forest to learn imbalanced data.
	\newblock University of California, Berkeley. 2004;.
	
	\bibitem{ling1998data}
	Ling CX, Li C.
	\newblock Data Mining for Direct Marketing: Problems and Solutions.
	\newblock In: KDD. vol.~98; 1998.  p. 73--79.
	
	\bibitem{drummond2003c4}
	Drummond C, Holte RC, et~al.
	\newblock {C4.5}, class imbalance, and cost sensitivity: why under-sampling
	beats over-sampling.
	\newblock In: Workshop on Learning from Imbalanced Datasets II. vol.~11.
	Citeseer; 2003. .
	
	\bibitem{kubat1997addressing}
	Kubat M, Matwin S, et~al.
	\newblock Addressing the curse of imbalanced training sets: one-sided
	selection.
	\newblock In: ICML. vol.~97. Nashville, USA; 1997.  p. 179--186.
	
	\bibitem{aslam2007estimating}
	Aslam JA, Popa RA, Rivest RL.
	\newblock On estimating the size and confidence of a statistical audit.
	\newblock In: USENIX/ACCURATE Electronic Voting Technology Workshop (EVT);
	2007.  p. 1--12.
	
	\bibitem{schafer1998multiple}
	Schafer JL, Olsen MK.
	\newblock Multiple imputation for multivariate missing-data problems: A data
	analyst's perspective.
	\newblock Multivariate behavioral research. 1998;33(4):545--571.
	
	\bibitem{zdrahal1981recognition}
	Zdrahal Z, Bratko I, Shapiro A.
	\newblock Recognition of complex patterns using cellular arrays.
	\newblock The Computer Journal. 1981;24(3):263--270.
	
\end{thebibliography}

\end{document}